\documentclass{ecai} 
\usepackage{latexsym}
\usepackage{amssymb}
\usepackage{amsmath}
\usepackage{amsthm}
\usepackage{booktabs}
\usepackage{enumitem}
\usepackage{graphicx}
\usepackage{color}
\usepackage{multirow}

\newcommand{\BibTeX}{B\kern-.05em{\sc i\kern-.025em b}\kern-.08em\TeX}

\begin{document}

%%%%%%%%%%%%%%%%%%%%%%%%%%%%%%%%%%%%%%%%%%%%%%%%%%%%%%%%%%%%%%%%%%%%%%%%

\begin{frontmatter}

%%% Use this command to specify your submission number.
%%% In doubleblind mode, it will be printed on the first page.

\paperid{1732} 

%%% Use this command to specify the title of your paper.

\title{Differentiating Choices via Commonality for Multiple-Choice Question Answering}

%%% Use this combinations of commands to specify all authors of your 
%%% paper. Use \fnms{} and \snm{} to indicate everyone's first names 
%%% and surname. This will help the publisher with indexing the 
%%% proceedings. Please use a reasonable approximation in case your 
%%% name does not neatly split into "first names" and "surname".
%%% Specifying your ORCID digital identifier is optional. 
%%% Use the \thanks{} command to indicate one or more corresponding 
%%% authors and their email address(es). If so desired, you can specify
%%% author contributions using the \footnote{} command.

%%%\author[A]{\fnms{Wenqing}~\snm{Deng}\orcid{....-....-....-....}\thanks{Corresponding Author. Email: somename@university.edu.}\footnote{Equal contribution.}}
%%%\author[B]{\fnms{Second}~\snm{Author}\orcid{....-....-....-....}\footnotemark}
%%%\author[B,C]{\fnms{Third}~\snm{Author}\orcid{....-....-....-....}}

\author[A]{\fnms{Wenqing}~\snm{Deng}}
\author[A]{\fnms{Zhe}~\snm{Wang}\thanks{Corresponding Author. Email: zhe.wang@griffith.edu.au.}}
\author[A]{\fnms{Kewen}~\snm{Wang}}
\author[A]{\fnms{Shirui}~\snm{Pan}}
\author[B]{\fnms{Xiaowang}~\snm{Zhang}}
\author[B]{\fnms{Zhiyong}~\snm{Feng}}

\address[A]{School of Information and Communication Technology, Griffith University,
Brisbane, Australia}
\address[B]{College of Intelligence and Computing, Tianjin University, Tianjin, China}

%%% Use this environment to include an abstract of your paper.

\begin{abstract}
Multiple-choice question answering (MCQA) becomes particularly challenging when all choices are relevant to the question and are semantically similar. Yet this setting of MCQA can potentially provide valuable clues for choosing the right answer. Existing models often rank each choice separately, overlooking the context provided by other choices. Specifically, they fail to leverage the semantic commonalities and nuances among the choices for reasoning. In this paper, we propose a novel MCQA model by differentiating choices through identifying and eliminating their commonality, called DCQA. Our model captures token-level attention of each choice to the question, and separates tokens of the question attended to by all the choices (i.e., commonalities) from those by individual choices (i.e., nuances). Using the nuances as refined contexts for the choices, our model can effectively differentiate choices with subtle differences and provide justifications for choosing the correct answer. We conduct comprehensive experiments across five commonly used MCQA benchmarks, demonstrating that DCQA consistently outperforms baseline models. Furthermore, our case study illustrates the effectiveness of the approach in directing the attention of the model to more differentiating features.
\end{abstract}

\end{frontmatter}

%%%%%%%%%%%%%%%%%%%%%%%%%%%%%%%%%%%%%%%%%%%%%%%%%%%%%%%%%%%%%%%%%%%%%%%%

\section{Introduction}

Multiple-choice question answering (MCQA) still presents a multifaceted challenge in natural language processing (NLP), especially when commonsense and reasoning are required \citep{huang2022clus, mihaylov2018knowledgeable,rajani2019explain, xu2020fusing}. 
The goal of MCQA is to enable machines to select from a set of choices the most relevant one to the question, and to make the task challenging, it often involves several distracting choices that are all relevant in various degrees and semantically similar to the answer. Commonsense and reasoning capabilities are required to differentiate the answer from distracting choices. Several datasets, such as CommonsenseQA \citep{talmor2019commonsenseqa} and OpenBookQA \citep{mihaylovCKS2018openbookqa}, have been created to aid researchers in developing and evaluating models for commonsense reasoning in MCQA.

Pre-trained language models (LMs) like GPT-3 \citep{brown2020language}, BERT \citep{devlin2019bert}, and RoBERTa \citep{liu2019roberta} demonstrate impressive reasoning capabilities and have led to significant progress in MCQA tasks. However, their reasoning processes are often considered implicit and lacking in explainability \citep{wang2023Dynamic}. Recent studies \citep{shen22an,yin22Interpreting} have emphasized the importance of explainability in NLP tasks, prompting research efforts to design MCQA models with interpretable reasoning processes.
Another line of research tries to incorporate external knowledge bases, such as knowledge graphs, to perform explicit reasoning and use such external knowledge as explanations \citep{bi2019incorporating,lv2020graph,taunk2023grapeqa,Yasunaga2022deep,zhang22greaselm}. While these approaches have demonstrated potential in improving interpretability, it's important to note that not all required commonsense knowledge can be captured in such an external knowledge base. Additionally, due to their large sizes, incorporating such knowledge bases would likely introduce a significant amount of noise.

\begin{figure}[t]
		  \begin{center}		  
    \includegraphics[width=0.9\columnwidth]{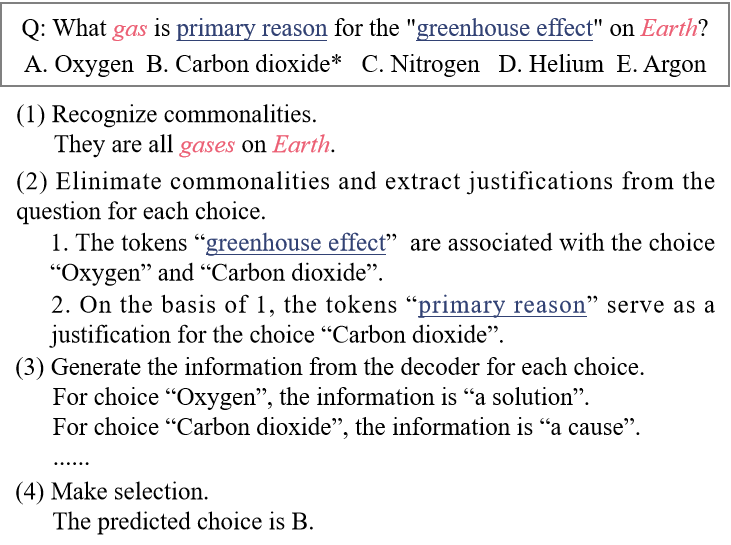}
		    \caption{An example of MCQA with its reasoning process to derive the correct choice B. The red tokens in the question are common to all choices and the blue tokens are differentiating ones for choosing the answer.}
            \label{Figure_1}
            \vspace{0.5cm}
            \end{center}
\end{figure}

Recent research has further underscored the pivotal role of leveraging the contextual information within the questions and choices for interpretable reasoning in MCQA tasks \citep{chen2023distinguish,huang2022clus, sun2019improving,zhang2020dcmn}. 
For instance, GenMC \citep{huang2022clus} delves into the contextual information within questions to generate \emph{clues} for selecting the answers, and such clues serve as intermediate steps of the reasoning. Nonetheless, such clues are generated independently of the choices and thus do not utilize contextual information among the choices. Indeed, the setting of MCQA can potentially provide valuable clues from the choices, including the distracting ones due to their semantic similarity to the answers. Yet existing MCQA models typically process each question-choice pair separately to assess each choice independently, overlooking the contextual information provided by other choices. Specifically, they fail to leverage the semantic commonalities and nuances among the choices for reasoning. In contrast, human beings often tackle MCQA by collectively comparing all choices and scrutinizing their differences to derive clues.

One form of contextual information often overlooked by independent reasoning is the commonalities among choices, such as shared themes or attributes. 
To illustrate this point, consider a scenario depicted in Figure \ref{Figure_1}, where all choices share a commonality of being gases on Earth. Existing MCQA models may associate all choices with the token ``gas'' in the question (reflected in high token-level attention weights), leading to potential confusion or misinformation. By prioritizing the identification and mitigation of such commonalities during the reasoning process, justifications or differentiating clues for each choice can be extracted from the question. For example, while both ``Oxygen'' and ''Carbon dioxide'' are associated with the tokens ''greenhouse effect'', the tokens ''primary reason'' serve as a justification for the choice ''Carbon dioxide''. By further generating explanatory information for the refined questions, we can obtain a distinguishing clue for each choice. For example, generating the information ''a cause'' for ''carbon dioxide'' reinforces the rationale for selecting ''Carbon dioxide''. Therefore, reasoning among the choices to identify commonalities, detect subtle differences, and generate differentiating clues is crucial for MCQA.

To address this limitation, we introduce a novel approach called Differentiating Choices via Commonality for MCQA (DCQA). Our approach first examines the choices to generate a representation of the commonalities shared among them. We then use this representation to derive a representation of the question tailored to each specific choice (called as a choice-specific representation of the question in this paper), effectively isolating the unique contextual information relevant to that choice. This choice-specific representation of the question serves as a clue or justification within the question for that particular choice. Subsequently, the decoder activates contextual information to the choice from the choice-specific representation of question. By integrating this information, we enhance the representation of each choice. In our experiments across five commonly used MCQA benchmarks, we demonstrate that our model consistently outperforms baseline methods, showcasing its ability to improve MCQA systems. Furthermore, a case study illustrates how our model effectively captures the commonalities among choices and identifies distinctive clues within questions, ultimately leading to enhanced MCQA performance. Our code is publicly available at \url{https://github.com/dwq-vicki/DCQA}.

%%%%%%%%%%%%%%%%%%%%%%%%%%%%%%%%%%%%%%%%%%%%%%%%%%%%%%%%%%%%%%%%%%%%%%%%

\section{Related Work}

Recently, large language models (LLMs) like GPT-3 \citep{brown2020language} and DeBERTa \citep{he2021deberta} have demonstrated impressive reasoning capabilities and have emerged as dominant approaches for multiple-choice question answering (MCQA) tasks. However, the use of larger LLMs often leads to disproportionate resource consumption and longer training times \citep{wang2023Dynamic}. Additionally, these models exhibit implicit reasoning processes and lack explainability, particularly in commonsense reasoning tasks. Consequently, several efforts have focused on enhancing machine reasoning abilities. In the following sections, we will provide a detailed overview of these efforts.

\subsection{Knowledge-Enhanced Models on MCQA Tasks} \label{RW_1}
Several models in MCQA have incorporated external knowledge sources to enhance the reasoning process and interpretability. These models utilize knowledge graphs (KGs) or other unstructured knowledge representations to augment the understanding of questions and choices. Some approaches focus on concatenating the commonsense knowledge with the question and encoding it into embeddings by using LMs with self-attention mechanisms \citep{chen2022dictbert,xu2020fusing,xu2022human}. Other methods leverage graph neural networks (GNNs) to integrate commonsense knowledge and perform reasoning within the graph structure, often referred to as KG-based approaches \citep{chen2020improving,feng2020scalable,lv2020graph,taunk2023grapeqa,wang2023Dynamic,yan2021learning,yasunaga2021qa,Yasunaga2022deep,zhang22greaselm}. 
For example, Chen et al. propose a graph-based iterative knowledge retrieval module, which iteratively retrieves concepts and entities related to the given question and its choices from multiple knowledge sources \citep{chen2020improving}.  MHGRN \citep{feng2020scalable} introduces a multi-hop relational reasoning module to perform multi-hop, multi-relational reasoning over subgraphs extracted from external knowledge graphs. GrapeQA \citep{taunk2023grapeqa} employs a graph augmentation method for prominent entities by identifying relevant text chunks from the QA pair and augments the working graph with corresponding latent representations from the LM. DHLK \citep{wang2023Dynamic} introduces a dynamic heterogeneous-graph reasoning method using LMs and knowledge representation learning. HGN \citep{yan2021learning} learns to jointly contextualize extracted and generated knowledge by reasoning over both within a unified graph structure to filter out context-irrelevant edges. QAGNN \citep{yasunaga2021qa} estimates the importance of subgraph entities using LMs and considers the QA context as an additional node connected to the subgraph, enabling joint reasoning over the QA context and KG. 

However, these methods often face limitations as not all commonsense knowledge can be captured in an external knowledge base. Furthermore, while KGs excel in explicit reasoning, they may struggle to capture implicit, context-dependent rationale \citep{Kawabata23Evaluating}. In contrast, our model takes a different approach by not relying on external knowledge sources. We capture only the contextual information within the QA task itself to facilitate reasoning, potentially mitigating some of the challenges faced by KG-based approaches.

\subsection{Contextual Information Models on MCQA Task}  \label{RW_2}
Numerous MCQA models have been developed to leverage contextual information within the questions and choices for commonsense reasoning. For instance, Sun et al. propose three reading strategies inspired by human behavior for extracting contextual information between questions and choices: back-and-forth reading, highlighting important information, and self-assessment through self-generated questions \citep{sun2019improving}. Zhang et al. introduce a two-way matching strategy for reading comprehension, to extract contextual information between questions and choices \citep{zhang2020dcmn}. There has been a growing interest in developing unified frameworks to address various tasks requiring contextual information in natural language understanding. One notable example is the UnifiedQA model \citep{khashabi2020unifiedqa}, which integrates 20 QA datasets into a unified training format and achieves state-of-the-art performance on various MCQA datasets. Similarly, GenMC \citep{huang2022clus} proposes a framework for generating contextual called as clues from questions to infer choices in MCQA, demonstrating impressive results on five different MCQA datasets and providing visualizations for interpretability.

However, these models typically perform reasoning or generate contextual information based on each question-choice pair independently, overlooking valuable context from the interaction among the choices. In our model, we identify and eliminate the commonalities among the choices to obtain the nuances for differentiating these choices. By focusing on the refined contexts for the choices, our model enhances the reasoning capabilities of MCQA systems.

%%%%%%%%%%%%%%%%%%%%%%%%%%%%%%%%%%%%%%%%%%%%%%%%%%%%%%%%%%%%%%%%%%%%%%%%

\begin{figure*}[ht]
        \centering
\includegraphics[width=1.5\columnwidth,height=7cm]{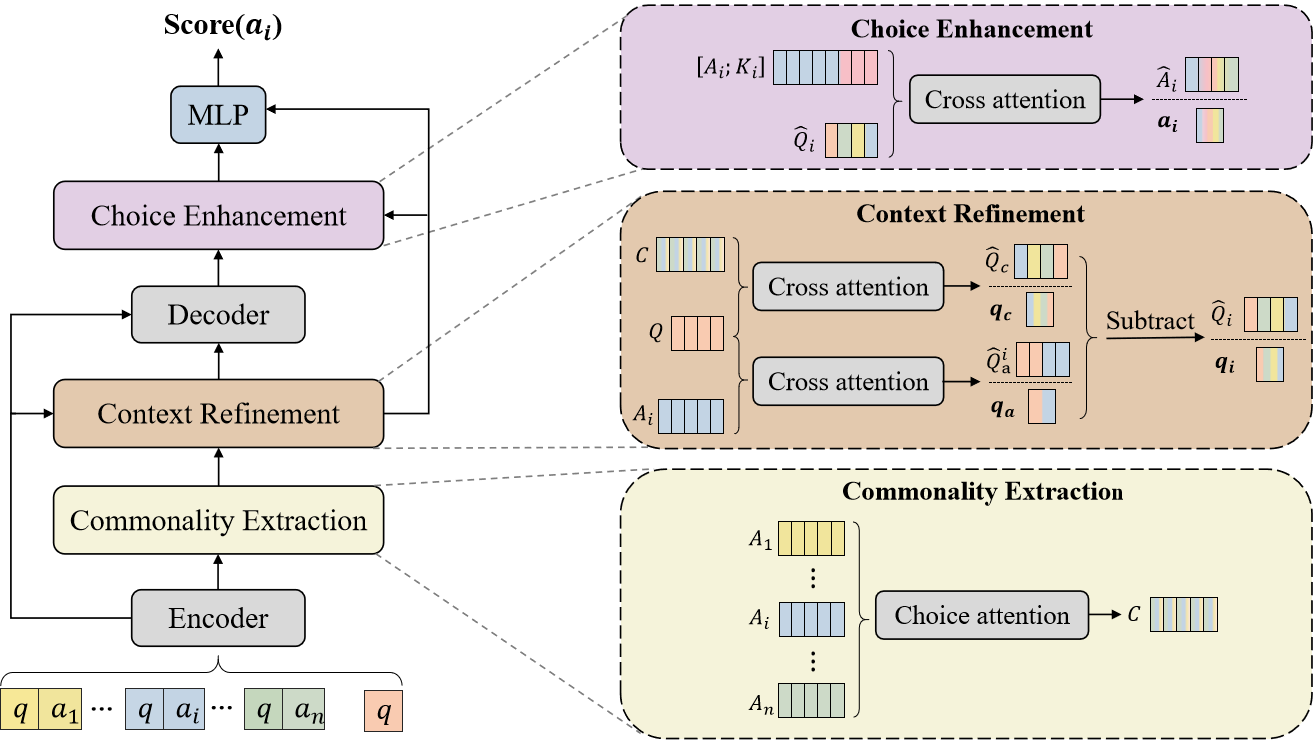}
		    \caption{Overall architecture of our proposed DCQA model. We input the Q context (question) as the representation of the question and the QA context (question + choice) as the representation of each choice.}
            \vspace{0.5cm}
		    \label{Figure_2}
      
\end{figure*}

\section{Method}
In this section, we formalize the MCQA problem and then introduce a new model for MCQA, call Differentiating Choices via Commonality for MCQA (DCQA), which emphasizes the differences between the choices via their commonalities.

Formally, given a natural language question $q$ 
and $n$ choices $a_{1}, a_{2}, \ldots, a_{n}$, 
the task is to determine the most plausible choice. Like most models for MCQA, this is typically achieved by effectively ranking the $n$ choices based on a confidence score for each choice w.r.t. the question. 

An overview of our model is illustrated in Figure \ref{Figure_2}. Our model comprises four primary modules. Firstly, the Context Representation module encompasses the initial embeddings of the questions and choices (Section \ref{CR}). Secondly, in view of the question, the Commonality Extraction module extracts a representation of the commonalities among all choices (Section \ref{CE}). Thirdly, the Context Refinement module is designed to obtain the refined context as nuances by generating a representation of the question tailored to each specific choice via the commonalities. This representation is referred to as a choice-specific representation of the question (Section \ref{UQ}). Finally, the Choice Enhancement module refines the representation of each choice using the corresponding refined context obtained from the previous module (Section \ref{UC}).

\subsection{Context Representation} \label{CR}
In line with the methodology of GenMC \citep{huang2022clus}, we adopt a similar approach for obtaining the embeddings of the questions and choices. We input the Q context (question) as the representation of the question and the QA context (question + choice) as the representation of each choice. To encode these representations, we employ a pre-trained encoder-decoder model, following the equations:
\begin{align}
Q & = \mathsf{encoder}(q), \\
A_i & = \mathsf{encoder}(q + a_i),
\end{align}
where $Q \in \mathbb{R}^{l\times d}$ and $A_i\in \mathbb{R}^{m\times d}$ ($1\le i\le n$). Here, $l$ and $m$ represent the maximum length of the question and choices in terms of tokens, and $d$ denotes the dimension of the encoding for each token.

We opt for an encoder-decoder model to capitalize on the natural language understanding capabilities of language models. This choice is driven by our aim to utilize the decoder to generate contextual information pertaining to the question, which is embedded within the LM. The encoder in our model captures the contextual information of the question and choices, while the decoder assists in generating supplementary contextual information, as will be discussed in the Choice Enhancement module (Section \ref{UC}).

\subsection{Commonality Extraction} \label{CE}

When tackling MCQA tasks, human beings often compare all choices simultaneously and analyze their differences to derive clues. This strategy allows humans to leverage the nuances and subtleties in the choices, enabling them to make more informed decisions. However, directly extracting differentiating clues for each choice without considering the other choices is challenging. Therefore, in this module, we first compare all choices to extract a representation of their commonalities $c$ among all choices $a_1, \ldots, a_n$. This representation is then used in the next module to extract differentiating clues for each choice based on the given question.

Inspired by \citep{zhang2020dcmn}, we propose an attention mechanism called choice-attention to calculate the commonalities $c$. In the previous module, the encoder encodes each choice $a_i$ as $A_i$. Firstly, we compute the interaction matrix between any two choices $a_i$ and $a_j$ by leveraging matrix multiplication. We then use $\mathsf{softmax}$ function to compute the similarity score matrix $S_{i,j}$ as follows:
\begin{align}
S_{i,j} &= \mathsf{softmax}(A_i W_{i,j} A_j^T),
\end{align}
where $1\le i,j\le n$, $S_{i,j}\in\mathbb{R}^{m\times m}$ and  $W_{i,j}\in\mathbb{R}^{d\times d}$ is a learnable parameter matrix. Each entry $s_{i,j}\in S_{i,j}$ denotes the similarity score between two tokens from choice $a_i$ and $a_j$, respectively.

Next, we extract the interaction matrix representation for each choice $a_i$ by weighting the sum of the similarity score matrix $S_{i,j}$ with other choices $A_j$. Here, we consider such representation as the commonalities matrix specific to choice $a_i$. In fact, each choice obtains a commonalities matrix from the other choices, resulting in $n$ commonalities matrices. By summing all the commonalities matrices and computing the average, we aggregate the commonalities among the choices to form a representation $C$ as follows:
\begin{align}
C & = (\sum_{1\le i\le n} \sum_{1\le j\le n,i\neq j} S_{i,j} A_j)/n, 
\end{align}
where $C\in\mathbb{R}^{m\times d}$ is the unified representation of  commonalities among all choices.

A key difference between our model and the inspiring work \citep{zhang2020dcmn} is that our model obtains a common embedding from all choices to differentiate them by eliminating such common information, while the latter obtains the interaction among all choices to enhance the representation of each choice.

\subsection{Context Refinement} \label{UQ}

After deriving the representation $C$ of the commonalities among all the choices, our model uses $C$ to generate a tailored representation of the question for each specific choice, referred to as a choice-specific representation of the question. As illustrated in Figure \ref{Figure_1}, to obtain this representation, we first identify tokens in the question related to the commonality of all choices and reduce their weights in representation. Subsequently, we identify the unique tokens as justifications for each choice and enhance their representation. In this module, we simulate this process by extracting two types of information: contextual information related to the commonalities $c$ and contextual information related to each choice $a_i$. 

\textbf{Contextual Information Related to the Commonalities.} Inspired by \citep{zhang2020dcmn}, we introduce a cross-attention mechanism to extract relevant contextual information from the question for the commonalities. This mechanism operates at the token level, assigning higher weights to more relevant tokens. We first compute two weight matrices, $I_q$ and $I_c$, between the question $q$ and the commonalities $c$. We then extract the representation $Q_c$ of interaction information from the commonalities $c$ by utilizing the weight matrix $I_c$. Finally, we incorporate this interaction information $Q_c$ into the question $q$ to get the representation $\widehat{Q}_c$ of contextual information in the question $q$ related to the commonalities $c$ as follows:
\begin{align}
I_q & = \mathsf{softmax}(CQ^T), \label{equation7}\\
I_c & = \mathsf{softmax}(QC^T), \\
Q_c & = I_c[I_qQ;C], \\
\widehat{Q}_c & = \mathsf{normal}(Q+Q_cW_I), \\
\mathbf{q_c} & = \mathsf{MaxPooling}(\widehat{Q}_c), 
\end{align}
where $I_q\in\mathbb{R}^{m\times l}$, $I_c\in\mathbb{R}^{l\times m}$, $Q_c\in\mathbb{R}^{l\times 2d}$, $\mathbf{q_c}\in\mathbb{R}^d$ and $W_I\in \mathbb{R}^{2d\times d}$ is a learnable parameter matrix. $[X;Y]$ denotes the concatenation of the matrices $X$ and $Y$. Here, $I_q$ represents the weight of each token in the question $q$ for the commonalities $c$, while $I_c$ represents the weight of each token in the commonalities $c$ for the question $q$. The vector $\mathbf{q_c}$ is an embedding that represents the entire semantic information contained in $\widehat{Q}$. In our model, $\mathbf{q_c}$ is used for the final computation of the score for each choice.

The equations described above constitute the entire cross-attention mechanism: $\widehat{Q}_c, \mathbf{q}c = \mathsf{att_{cross}}(Q;C)$.

\textbf{Contextual Information Related to Each Choice.}
We employ the same cross-attention mechanism to calculate the contextual information from the question $q$ related to each choice $a_i$: $\widehat{Q}_a^i, \mathbf{q}_a^i = \mathsf{att_{cross}}(Q;A_i)$, where $1\le i\le n$.

To derive the choice-specific representation of the question as refined context, we subtract the embedding $\widehat{Q}_c$ from $\widehat{Q}_a^i$, simulating the process of eliminating the commonalities information in the question for each choice. Similar operation is conducted on the entire semantic information between $\mathbf{q}_c$ and $\mathbf{q}_a^i$:
\begin{align}
\widehat{Q}_i & = \widehat{Q}_a^i - \widehat{Q}_c, \\
\mathbf{q}_i & = \mathbf{q}_a^i - \mathbf{q}_c,
\end{align}

\subsection{Choice Enhancement} \label{UC}

The choice-specific representation of the question, obtained from the previous module, offers a refined contextual understanding of the question for each choice. As depicted in Figure \ref{Figure_1}, we can further generate the information from the choice-specific representation of the question to provide distinct contextual information for each choice. Therefore, in this module, we first leverage the decoder to generate contextual information $K_i$ related to the choice-specific representation of the question, as follows:
\begin{align}
K_i & = \mathsf{decoder}(\widehat{Q}_i), 
\end{align}
Here, $K_i\in \mathbb{R}^{p\times d}$, where $p$ and $d$ denote the length of tokens in $K_i$ and the representation dimension, respectively.

We use the original QA context as the semantic representation of each choice in Section \ref{CR}, but we have refined the semantic information of the question for each choice. Therefore, we need to enhance the representation of each choice using the choice-specific representation $\widehat{Q}_i$ of the question and the generated contextual information $K_i$. Similar to the Context Refinement module (Section \ref{UQ}), we employ the cross-attention mechanism to calculate the enhanced representation $\widehat{A}_i$ and $\mathbf{a}_i$ for each choice: $\widehat{A}_i, \mathbf{a}_i = \mathsf{att_{cross}}([A_i;K_i], \widehat{Q}_i)$, where $\widehat{A}_i\in \mathbb{R}^{(m+p)\times d}$ and $\mathbf{a}_i\in \mathbb{R}^d$.

We compute the confidence score of each choice $a_i$ based on the semantic connection between $\mathbf{q}_i$ and $\mathbf{a}_i$. This score is calculated using an $\mathsf{MLP}$ and $\mathsf{softmax}$ layer as follows:
\begin{equation}
    score(a_i) = \mathsf{softmax}\left(\mathsf{MLP}\left[\mathbf{q}_i; \mathbf{a}_i\right]\right).
\end{equation}
Finally, we select the choice with the highest score as the predicted choice. 

%%%%%%%%%%%%%%%%%%%%%%%%%%%%%%%%%%%%%%%%%%%%%%%%%%%%%%%%%%%%%%%%%%%%%%%%

\section{Experiments}
\subsection{Datasets} \label{Datasets}

We evaluated our model on five widely used commonsense MCQA benchmarks. The CommonsenseQA (CSQA) \citep{talmor2019commonsenseqa} and OpenBookQA (OBQA) \citep{mihaylovCKS2018openbookqa} datasets are widely used commonsense QA benchmarks and constructed by crowed-sourcing. The ARC-Easy (ARC-E) and ARC-Challenge (ARC-C) datasets are subsets of the AI2 Reasoning Challenge (ARC) benchmark \citep{clark2018think}, which focuses on scientific questions. The QA via Sentence Composition (QASC) dataset \citep{khot2020qasc} is a collection of (primary and middle) school-level science questions. Since the correct answers of the official Test sets are not released for the CSQA and QASC benchmarks, we used their official dev set as our test set for experiments. Additionally, we randomly held out an in-house dev set from the training set, as described in \citep{huang2022clus}.
\begin{table}[ht]
	\centering
        \caption{Statistics of the benchmark datasets.}
        \vspace{0.5cm}
	%\small
        \resizebox{76mm}{12mm}{
	\begin{tabular}{lcccccc}
	\hline
	  & {Train} & {Dev} & {Test} & {\#C} & {|Q|} & {|C|}\\
	\hline
	CSQA & 8500 & 1241 & 1221 & 5 & 13.4 & 1.5\\
	OBQA & 4957 & 500 & 500 & 4 & 10.7 & 2.9\\
	ARC-E & 2241 & 567 & 2365 & 4 & 19.4 & 3.7\\
	ARC-C & 1117 & 295 & 1165 & 4 & 22.3 & 4.9\\
	QASC & 7320 & 814 & 926 & 8 & 8.1 & 1.6\\
	\hline
	\end{tabular}}
 \label{tab:benchmarks}
\end{table}

The statistics of the benchmarks are in Table \ref{tab:benchmarks}, where we record for each dataset, the sizes of the Train, Dev, and Test sets, the number of choices (\#C), and the average lengths of questions (|Q|) and choices (|C|).

Additional Datasets and their experimental results can be found in the Appendix \ref{additional_datasets}.
\subsection{Experimental Setup} \label{Experimental Setup}

To ensure a fair comparison with baselines, we selected two popular encoder-decoder language models as a basis, T5 \citep{wolf2019huggingface} and Unifiedqa-T5 \citep{khashabi2020unifiedqa}. We conducted experiments under their Base and Large versions. In our experiments, we set the maximum length of the input sequence to 64 and the output embedding dimension to 1024. We employed PyTorch 1.8 framework and utilized the Adam optimizer \citep{loshchilov2019decoupled} with a weight decay rate of 0.01. The number of training iterations was set to 50, with early stopping after 15 epochs if no improvement in accuracy on the dev  dataset was observed. To determine the optimal learning rate and batch size, we conducted a search over the ranges \{1e-4, 5e-5, 1e-5, 5e-6\} and \{16, 8, 4\}, respectively, which are shown in Table \ref{tab:parameters}.

\begin{table}[ht]
	\centering
        \caption{The selection of the optimal learning rate and batch size.}
        \vspace{0.5cm}
	%\small
        \resizebox{85mm}{!}{
	\begin{tabular}{lcccccc}
	\hline
	  Models & Hyperparameters & {CSQA} & {OBQA} & {ARC-E} & {ARC-C} & {QASC} \\
	\hline
	\multirow{2}{*}{{T5}} & Learning Rate & 1e-5 & 5e-5 & 1e-4 & 1e-4 & 5e-5\\
	& Batch Size & 16 & 16 & 4 & 16 & 8\\
        \hline
       \multirow{2}{*}{{Unified-T5}} & Learning Rate & 1e-4 & 1e-4 & 1e-4 & 1e-4 & 1e-4\\
	& Batch Size & 8 & 8 & 16 & 8 & 8\\
	\hline
	\end{tabular}
        }
 \label{tab:parameters}
\end{table}

Considering that neural models are sensitive to different random seeds, we performed multiple experiments for each model with different random seeds, and reported the mean and standard deviation. For CSQA, OBQA, ARC-Easy, and QASC, we used three random seeds \{1,10,20\}. For the smallest dataset ARC-Challenge, we used five random seeds \{1,10,20,30,40\} to ensure the authenticity and accuracy of our results. All experiments were conducted on an NVIDIA PH402 SKU 200 with 32G memory.

\subsection{Baselines}
Our model aims to differentiate choices by using contextual information that represents the commonalities among all choices, without relying on external knowledge bases. To align with this design, we compare our model with several state-of-the-art MCQA models that solely utilize encoder-decoder architecture, excluding external knowledge. These models include GenMC \citep{huang2022clus} and UnifiedQA \citep{khashabi2020unifiedqa}, which were discussed in Section \ref{RW_2}. Additionally, we compare with T5-vanilla \citep{Colin20Exploring}, as in GenMC. Specifically, we concatenate the question with all choices, with each choice preceded by its respective choice ID. The entire sequence is then prepended with the dataset name. This concatenated sequence is input into the encoder, which generates a joint representation of the question and all choices. Based on this joint representation, the decoder outputs the corresponding choice ID. In this configuration, the decoder essentially functions as a classifier. Our comparison focuses on accuracy, evaluating how effectively each model selects the correct choice.

\subsection{Main Results and Analysis} \label{main_results}
\begin{table*}[ht]
    \begin{center}
    \caption{Comparison on MCQA benchmarks (CSQA, OBQA, ARC-E, ARC-C, QASC). All results are from our evaluations.}
    \vspace{0.3cm}
	%\small
    \resizebox{\textwidth}{!}{
	\begin{tabular}{lcccccccccc}
	\hline
	\multirow{2}{*}{{Model}} & \multicolumn{2}{c}{{CSQA}}  & \multicolumn{2}{c}{{OBQA}} & \multicolumn{2}{c}{{ARC-E}} & \multicolumn{2}{c}{{ARC-C}} & \multicolumn{2}{c}{{QASC}}\\
	& {Dev} & {Test} & {Dev} & {Test} & {Dev} & {Test} & {Dev} & {Test} & {Dev} & {Test}\\
	\hline
	T5-Base\\
        \quad T5-vanilla & 56.59($\pm$0.31) & 60.42($\pm$0.54) & 59.93($\pm$1.16) & 56.60($\pm$1.70) & 49.62($\pm$1.17) & 50.75($\pm$1.04) & 29.63($\pm$1.02) & 28.19($\pm$1.64) & 53.28($\pm$0.47) & 31.64($\pm$1.00)\\
        \quad GenMC\textsubscript{T5} & 61.05($\pm$0.40) & \textbf{62.87($\pm$0.22)} & \textbf{63.00($\pm$0.33)} & 61.27($\pm$1.60) & 60.43($\pm$0.42) & 56.27($\pm$0.61) & 38.51($\pm$0.90) & 36.53($\pm$0.40) & \textbf{59.67($\pm$0.65)} & 41.76($\pm$0.51)\\
	\quad Our Model & \textbf{61.56($\pm$0.34)} & 62.41($\pm$0.24) & 62.40($\pm$0.33) & \textbf{61.87($\pm$0.62)} & \textbf{61.49($\pm$0.68)} & \textbf{57.17($\pm$0.60)} & \textbf{38.58($\pm$0.94)} & \textbf{37.05($\pm$0.48)} & 59.50($\pm$0.90) & \textbf{42.98($\pm$1.57)}\\
        \hline
        U-T5-Base\\
	\quad UnifiedQA\textsubscript{T5} & 42.14($\pm$0.00) & 45.05($\pm$0.00) & 59.00($\pm$0.00) & 55.40($\pm$0.00) & 51.32($\pm$0.00) & 49.09($\pm$0.00) & 43.73($\pm$0.00) & 39.57($\pm$0.00)  & 15.97($\pm$0.00) & 24.41($\pm$0.00)\\
	\quad UnifiedQA\textsubscript{T5-FT} & 56.76($\pm$0.80) & 59.60($\pm$0.74) & 63.00($\pm$0.33) & 58.67($\pm$0.53) & 55.85($\pm$0.42) & 56.19($\pm$0.30) & 44.20($\pm$0.90) & \textbf{41.68($\pm$0.81)}  & 55.41($\pm$1.50) & 38.48($\pm$0.59)\\
        \quad Our Model & \textbf{61.00($\pm$0.64)} & \textbf{63.64($\pm$0.31)} & \textbf{64.07($\pm$0.34)} & \textbf{62.0($\pm$0.99)} & \textbf{61.96($\pm$0.54)} & \textbf{58.52($\pm$0.31)} & \textbf{46.44($\pm$1.08)} & 40.24($\pm$0.60) & \textbf{60.28($\pm$0.35)} & \textbf{44.74($\pm$0.42)}\\
	\hline
	T5-Large\\
        \quad T5-vanilla & 67.26($\pm$0.30) & 70.54($\pm$0.78) & 65.20($\pm$0.65) & 63.13($\pm$1.09) & 61.67($\pm$1.09) & 59.07($\pm$0.39) & 36.14($\pm$1.26) & 35.35($\pm$0.64) & 61.30($\pm$0.67) & 46.90($\pm$1.81)\\
        \quad GenMC\textsubscript{T5} & 69.75($\pm$0.04) & 72.42($\pm$1.35) & 69.10($\pm$0.82) & 67.40($\pm$0.75) & 70.49($\pm$0.30) & 66.55($\pm$0.69) & 45.49($\pm$1.84) & 44.58($\pm$1.74) & 67.20($\pm$1.39) & 54.75($\pm$3.58)\\
	\quad Our Model & \textbf{71.04($\pm$0.30)} & \textbf{73.05($\pm$0.27)} & \textbf{71.20($\pm$0.43)} & \textbf{67.70($\pm$0.47)} & \textbf{71.61($\pm$0.63)} & \textbf{67.20($\pm$0.17)} & \textbf{46.24($\pm$1.26)} & \textbf{45.99($\pm$0.62)} & \textbf{67.77($\pm$0.67)} & \textbf{58.17($\pm$0.19)}\\
        \hline
	U-T5-Large\\
        \quad UnifiedQA\textsubscript{T5} & 56.00($\pm$0.00) & 60.85($\pm$0.00) & 67.40($\pm$0.00) & 67.80($\pm$0.00) & 65.61($\pm$0.00) & 62.66($\pm$0.00) & 54.24($\pm$0.00) & 51.76($\pm$0.00) & 27.52($\pm$0.00) & 41.25($\pm$0.00)\\
	\quad UnifiedQA\textsubscript{T5-FT} & 68.85($\pm$0.04) & 71.88($\pm$0.54) & 71.33($\pm$0.68) & 68.60($\pm$0.16) & 71.45($\pm$0.27) & 67.56($\pm$0.14) & 52.54($\pm$0.74) & 51.81$\pm$(0.67) & 64.78($\pm$0.06) & 58.10($\pm$0.40)\\
	\quad Our Model & \textbf{71.45($\pm$0.68)} & \textbf{75.10($\pm$0.70)} & \textbf{72.73($\pm$1.20)} & \textbf{70.07($\pm$1.36)} & \textbf{73.72($\pm$0.50)} & \textbf{69.96($\pm$0.35)} & \textbf{52.54($\pm$0.96)} & \textbf{52.14($\pm$0.66)} & \textbf{68.26($\pm$0.42)} & \textbf{59.39($\pm$0.85)}\\
	\hline
	\end{tabular}
    }
    \label{tab:main_result}
    \end{center}
\end{table*}
Table \ref{tab:main_result} presents the comparison results on all five datasets. We categorize the results according to the language encoder, T5 and Unified-T5. Overall, our model demonstrates competitive performance with the baselines on both the Dev and Test sets of all datasets. Under the T5-Large configuration, our model achieves an improvement of 3.78\% and 2.51\% on the Dev and Test sets of CSQA over T5-vanilla, and surpasses GenMC\textsubscript{T5} by 1.29\% and 0.63\% on these sets. These results indicate the effectiveness of our reasoning method, particularly in MCQA tasks that require commonsense reasoning. When evaluated under the Unified-T5 setup, our model consistently outperforms the language model UnifiedQA\textsubscript{T5-FT} on all datasets except for the Test set of the ARC-C dataset. This exception may be attributed to the dataset's question and choice lengths, which introduce additional complexity. Specifically, with UnifiedQA\textsubscript{T5} as the base model, our model achieves improvements of 4.24\% and 4.04\% on the Dev and Test sets of the CSQA dataset, and 4.87\% and 6.26\% on the Dev and Test sets of the QASC dataset, respectively. However, our model's performance on the OBQA dataset with T5-Base is not as outstanding as GenMC, likely due to the relatively loose semantic connections among choices for each question. Additionally, the QASC dataset presents challenges for our model due to its large number of choices compared to other datasets, leading to some choices focusing on the same tokens and failing to be distinguished. In summary, our model exhibits strong performance across various MCQA benchmarks, showcasing its effectiveness in capturing and utilizing contextual information to differentiate choices.

\subsection{Ablation Study} \label{AS}
\begin{table*}[ht]
        \begin{center}
        \caption{Ablation study on MCQA benchmarks (CSQA, OBQA, ARC-E, ARC-C, QASC). The second row and the fourth row stand for the effectiveness of each module. The third row and the fifth row stand for the effectiveness of each cross-attention, respectively. All results are from our evaluations.}
        \vspace{0.3cm}
	%\small
    \resizebox{\textwidth}{!}{
	\begin{tabular}{lcccccccccc}
	\hline
	\multirow{2}{*}{{Model}} & \multicolumn{2}{c}{{CSQA}}  & \multicolumn{2}{c}{{OBQA}} & \multicolumn{2}{c}{{ARC-E}} & \multicolumn{2}{c}{{ARC-C}}  & \multicolumn{2}{c}{{QASC}}\\
	& {Dev} & {Test} & {Dev} & {Test} & {Dev} & {Test} & {Dev} & {Test} & {Dev} & {Test}\\
	\hline
	T5-Base\\
        \quad Our Model & \textbf{61.56($\pm$0.34)} & \textbf{62.41($\pm$0.24)} & \textbf{62.40($\pm$0.33)} & \textbf{61.87($\pm$0.62)} & \textbf{61.49($\pm$0.68)} & \textbf{57.17($\pm$0.60)} & \textbf{38.58($\pm$0.94)} & \textbf{37.05($\pm$0.48)} & 59.50($\pm$0.90) & \textbf{42.98($\pm$1.57)} \\
        \quad -ComE & 60.80($\pm$0.48) & 62.13($\pm$0.40) & 62.33($\pm$0.38) & 60.33($\pm$0.94) & 60.26($\pm$0.83) & 56.91($\pm$1.17) & 38.44($\pm$0.93) & 35.65($\pm$0.64) & \textbf{60.56($\pm$0.87)} & 42.12($\pm$1.17)\\
	\quad -CR & 60.89(($\pm$0.36) & 61.78(($\pm$0.57) & 62.00$\pm$(0.75) & 59.73($\pm$0.41) & 60.08($\pm$0.60) & 57.14($\pm$0.56) & 38.47($\pm$1.61) & 36.87($\pm$0.71) & 59.79($\pm$0.35) & 41.61($\pm$1.02)\\
        \quad -DE & 60.92($\pm$0.20) & 62.15($\pm$0.48) & 62.33($\pm$0.96) & 61.20($\pm$1.34) & 60.14($\pm$0.43) & 56.98($\pm$0.50) & 37.56($\pm$0.87) & 36.50($\pm$0.94) & 60.16($\pm$0.41) & 41.15($\pm$1.79)\\
        \quad -CE & 60.84($\pm$0.63) & 61.37($\pm$0.80) & 62.27($\pm$0.34) & 60.67($\pm$1.95) & 60.55($\pm$0.68) & 56.54($\pm$0.82) & 37.49($\pm$0.87) & 36.62($\pm$0.44) & 60.11($\pm$0.06) & 42.80($\pm$0.60)\\
	\hline
	T5-Base\\
        \quad Our Model & \textbf{61.56($\pm$0.34)} & \textbf{62.41($\pm$0.24)} & 62.40($\pm$0.33) & \textbf{61.87($\pm$0.62)} & \textbf{61.49($\pm$0.68)} & \textbf{57.17($\pm$0.60)} & 38.58($\pm$0.94) & \textbf{37.05($\pm$0.48)} & 59.50($\pm$0.90) & \textbf{42.90($\pm$1.57)} \\
        \quad -C1 & 60.80($\pm$0.48) & 62.13($\pm$0.40) & 62.33($\pm$0.38) & 60.33($\pm$0.94) & 60.26($\pm$0.83) & 56.91($\pm$1.17) & 38.44($\pm$0.93) & 35.65($\pm$0.64) & \textbf{60.56($\pm$0.87)} & 42.12($\pm$1.17)\\
	\quad -C2 & 60.89(($\pm$0.50) & 61.94(($\pm$0.84) & \textbf{62.93$\pm$(1.06)} & 59.60($\pm$1.82) & 60.67($\pm$0.00) & 56.21($\pm$0.10) & \textbf{38.98($\pm$1.77)} & 36.10($\pm$0.84) & 60.77($\pm$0.57) & 41.86($\pm$0.87)\\
        \quad -C3 & 60.84($\pm$0.63) & 61.37($\pm$0.80) & 62.27($\pm$0.34) & 60.67($\pm$1.95) & 60.55($\pm$0.68) & 56.54($\pm$0.82) & 37.49($\pm$0.87) & 36.62($\pm$0.44) & 60.11($\pm$0.06) & 42.80($\pm$0.60)\\
	\hline
T5-Large\\
        \quad Our Model & \textbf{71.04($\pm$0.30)} & \textbf{73.05($\pm$0.27)} & \textbf{71.2($\pm$0.43)} & \textbf{67.7($\pm$0.47)} & \textbf{71.61($\pm$0.63)} & 67.20($\pm$0.17) & \textbf{46.24($\pm$1.26)} & \textbf{45.99($\pm$0.62)} & 67.77($\pm$0.67) & 58.17($\pm$0.19)\\
        \quad -ComE & 70.91($\pm$0.20) & 72.07($\pm$0.52) & 69.70($\pm$0.19) & 67.10($\pm$0.67) & 70.78($\pm$0.58) & 66.51($\pm$0.67) & 46.17($\pm$0.78) & 45.73($\pm$1.14) & 67.12($\pm$0.60) & 57.17($\pm$0.79)\\
	\quad -CR & 70.94($\pm$0.42) & 72.54($\pm$0.22) & 70.33($\pm$0.50) & 66.80($\pm$0.86) & 70.96($\pm$0.55) & 67.10($\pm$0.48) & 45.15($\pm$0.25) & 44.29($\pm$1.26) & \textbf{68.47($\pm$0.12)} & 58.67($\pm$0.08)\\
        \quad -DE & 70.67($\pm$0.80) & 72.13($\pm$0.97) & 70.07($\pm$1.68) & 66.73($\pm$1.68) & 70.31($\pm$)1.10 & 66.71($\pm$0.35) & 45.96($\pm$1.35) & 45.51($\pm$0.67) & 68.02($\pm$0.55) & \textbf{59.12($\pm$1.31)}\\
        \quad -CE & 70.48($\pm$0.20) & 72.81($\pm$0.72) & 69.13($\pm$0.41) & 66.67($\pm$1.43) & 70.25($\pm$0.46) & \textbf{67.51($\pm$0.23)} & 45.83($\pm$0.78) & 45.39($\pm$1.53) & 67.73($\pm$0.55) & 55.83($\pm$1.78)\\
	\hline
        T5-Large\\
        \quad Our Model & \textbf{71.04($\pm$0.30)} & \textbf{73.05($\pm$0.27)} & \textbf{71.2($\pm$0.43)} & \textbf{67.7($\pm$0.47)} & \textbf{71.61($\pm$0.63)} & 67.20($\pm$0.17) & \textbf{46.24($\pm$1.26)} & \textbf{45.99($\pm$0.62)} & 67.77($\pm$0.67) & 58.17($\pm$0.19)\\
        \quad -C1 & 70.91($\pm$0.20) & 72.07($\pm$0.52) & 69.7($\pm$0.19) & 67.1($\pm$0.67) & 70.78($\pm$0.58) & 66.51($\pm$0.67) & 46.17($\pm$0.78) & 45.73($\pm$1.14) & 67.12($\pm$0.60) & 57.17($\pm$0.79)\\
	\quad -C2 & 70.83($\pm$0.24) & 72.67($\pm$0.67) & 69.33($\pm$0.90) & 65.93($\pm$1.80) & 69.54($\pm$0.22) & 65.64($\pm$1.42) & 42.64($\pm$1.49) & 44.29($\pm$1.80) & \textbf{68.39($\pm$0.59)} & \textbf{59.54($\pm$0.31)}\\
        \quad -C3 & 70.48($\pm$0.20) & 72.81($\pm$0.72) & 69.13($\pm$0.41) & 66.67($\pm$1.43) & 70.25($\pm$0.46) & \textbf{67.51($\pm$0.23)} & 45.83($\pm$0.78) & 45.39($\pm$1.53) & 67.73($\pm$0.55) & 55.83($\pm$1.78)\\
	\hline
	\end{tabular}
    }
    \label{tab:ablation_study_1}
    \end{center}
\end{table*}
To assess the contribution of each module and the three cross-attention methods, we conducted an ablation study. Table \ref{tab:ablation_study_1} summarizes the results, where C1, C2, and C3 denote the cross-attention mechanisms for the commonality feature of all choices and the question, the individual choice and the question, and the enhanced question concatenated with generated information, respectively.

The results indicate that, except for the Test set of ARC-E and the QASC dataset, removing the four variants (Commonality Extraction (ComE), Context Refinement (CR), Decoder (DE), and Choice Enhancement (CE)) results in a decrease in accuracy across all five datasets. Similarly, accuracy drops when we remove the three cross-attention mechanisms. This highlights the necessity of these modules and the essential role of the three cross-attention methods in the model.

However, on the QASC dataset, our model does not achieve the highest accuracy when considering the four variants and the three cross-attention methods. This discrepancy is likely due to the dataset's characteristics, such as having a larger number of choices and relatively shorter question lengths. These factors can result in similar information from the question regarding the choices, posing challenges in accurately distinguishing between them.

\subsection{Case Study}

This section showcases how our model effectively reduces the weights of commonalities compared to other models (Section \ref{Compared w Other Models}) and demonstrates the progressive enhancement of justifications within the question for each choice (Section \ref{Representation Refinement}).

To obtain these weights, we utilized an equation to generate a weight matrix between the question $Q$ and choice $A_i$: $\mathsf{softmax}(QA_i^T)$. We then applied MaxPooling to capture the attention scores of each token in $Q$ for each choice $A_i$. Since the distribution of these scores was relatively scattered, we applied Min-Max Normalization to scale these scores for each choice. Finally, we computed the percentage of each token in $Q$ based on these scaled scores.

\subsubsection{Attention Refinement}\label{Compared w Other Models}
\begin{figure*}[ht]
		  \centering
		  \includegraphics[width=2.0\columnwidth] {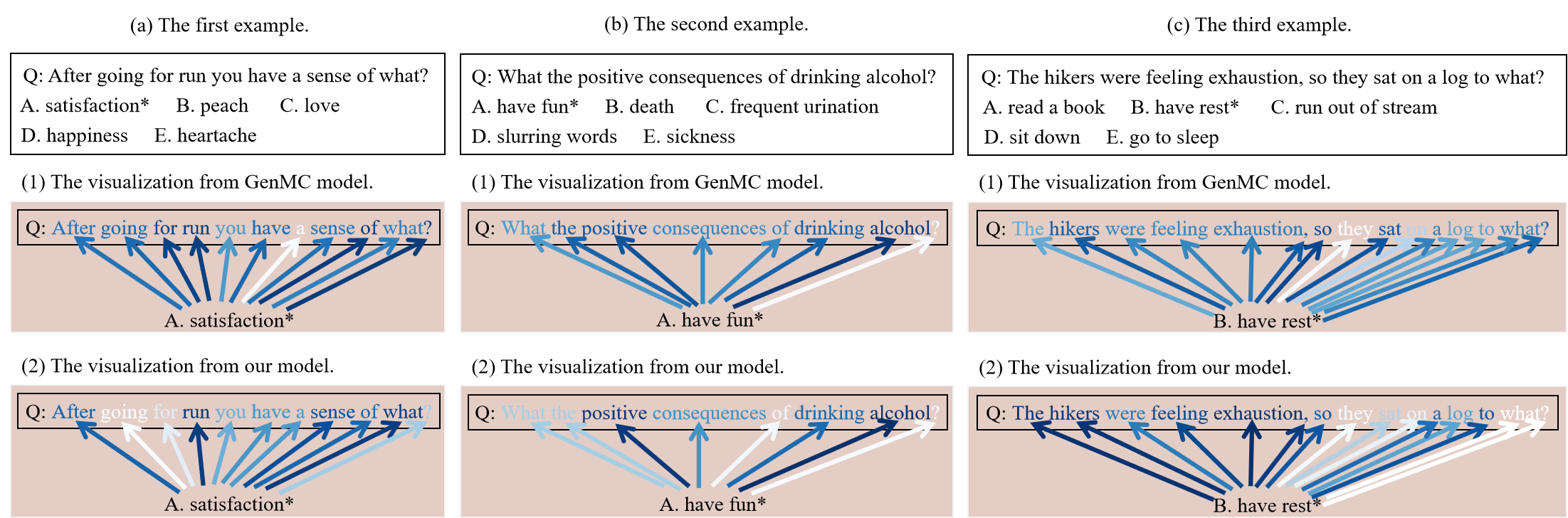}
            \vspace{0.3cm}
		    \caption{The comparison of weights visualization between the GenMC model and our model on three examples from the Dev dataset of CSQA. \newline The correct choice is marked with *. The darker the color is, the higher the weight is.}
		    \label{Figure_3}
\end{figure*}
Figure \ref{Figure_3} illustrates the effectiveness of eliminating commonalities, showcasing a comparative analysis with the baseline model GenMC. In example (a), GenMC assigns comparable weights to tokens such as ``\textit{what}'', ``\textit{for}'', and ``\textit{?}'' for the correct choice. In contrast, our model reduces the weights on these common tokens by eliminating commonalities, thus increasing the emphasis on tokens like ``\textit{run}'' and ``\textit{sense of what}''. Additionally, the token ``\textit{After}'' gains prominence, indicating its significance in the context of chronological order related to running. This highlights our model's ability to identify clues or justifications within the question for each choice, surpassing the capabilities of the GenMC model.

This efficacy is further exemplified in example (b), where our model accurately captures the critical token ``\textit{positive}''. Moreover, example (c) showcases our model's adeptness in handling lengthy sentences, further underscoring its robust performance.

\subsubsection{Representation Refinement} \label{Representation Refinement}
\begin{figure*}[ht]
		 \centering
   \includegraphics[width=2.0\columnwidth]{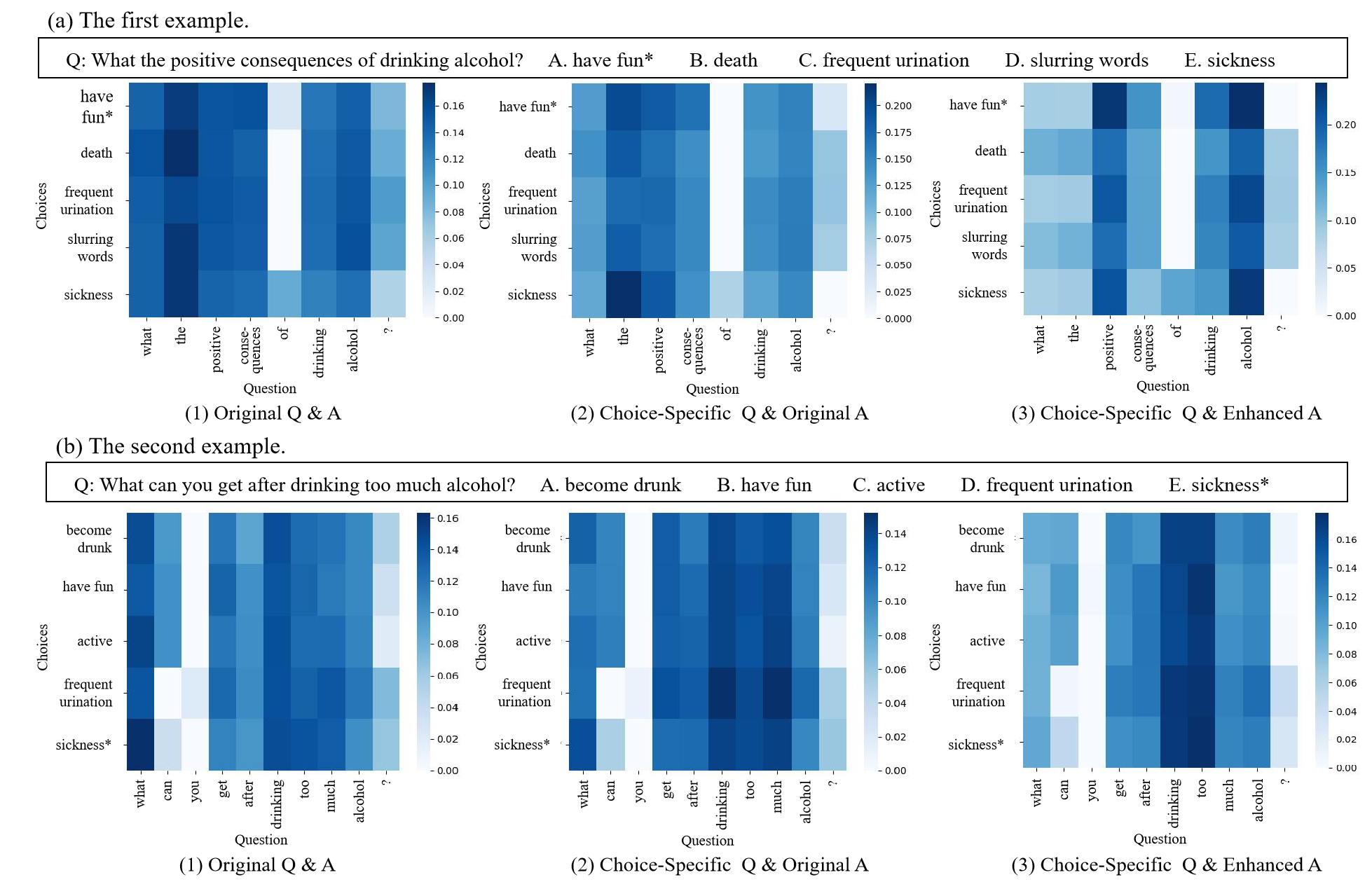}
		    \caption{The heatmaps of two examples from the Dev dataset of CSQA.}
      \vspace{0.2cm}
		    \label{Figure_4}
\end{figure*}

In this section, we illustrate the progressive refinement of question and choice representations in our model. Due to the separate generation of the choice-specific question representations and the enhanced choices, each example is accompanied by three heat maps. Figure \ref{Figure_4} presents these heat maps, where from left to right, the first heat map represents the attention between the original $Q$ and $A$ from the encoder, the second heat map represents the attention scores between the choice-specific representation of question $Q$ and the original $A$, and the third heat map signifies the interaction between the choice-specific representation of question $Q$ and the enhanced $A$.

In example (a), we observe that for the correct choice ``\textit{have fun}'', our model assigns higher weights to the tokens ``\textit{positive}'', ``\textit{driving}'', and ``\textit{alcohol}'' during the representation refinement process. Concurrently, the weights assigned to commonalities such as ``\textit{where}'' and ``\textit{the}'' gradually decrease.
Example (b) demonstrates a similar trend, where the weights assigned to commonalities ``\textit{what}'', ``\textit{can}'', ``\textit{you}'', and ``\textit{?}'' diminish. Both the correct choice ``\textit{sickness}'' and the incorrect choice ``\textit{frequent urination}'' receive attention on tokens ``\textit{drinking}'' and ``\textit{too}''. However, the correct choice emphasizes ``\textit{too}'', highlighting the concept of excessive drinking, which is crucial for distinguishing it from the incorrect choice.
These examples highlight our model's ability to recognize and leverage subtle differences among choices, showcasing its effectiveness in improving the accuracy of choice selection.

\subsection{Parameter Comparison} \label{PC}
We calculated the number of parameters for our model and the GenMC model under two configurations: T5-Base and T5-Large. In the T5-Base configuration, our model has 886.34 million parameters compared to GenMC's 895.37 million. Under the T5-Large configuration, our model has 2.81 billion parameters, while GenMC has 2.82 billion. Despite having fewer parameters, our model achieves higher accuracy than the GenMC model.

%%%%%%%%%%%%%%%%%%%%%%%%%%%%%%%%%%%%%%%%%%%%%%%%%%%%%%%%%%%%%%%%%%%%%%%%

\section{Conclusion}

In this paper, we introduce DCQA, a model designed for Multiple-Choice Question Answering (MCQA). Unlike traditional approaches that start from the question, DCQA uniquely begins with the choices, allowing it to extract relevant phrases from the question that are most pertinent to each individual choice. By utilizing token-level attention to separate tokens of the question attended to by all the choices (i.e., commonalities) from those attended to by individual choices (i.e., nuances), DCQA generates a choice-specific representation of the question, thereby facilitating the selection of the correct choice. Our experimental results demonstrate that DCQA outperforms baseline models across a range of datasets. Furthermore, our case study illustrates how our model effectively highlights the nuances of the choices, showcasing its interpretability and reasoning capabilities.

%%%%%%%%%%%%%%%%%%%%%%%%%%%%%%%%%%%%%%%%%%%%%%%%%%%%%%%%%%%%%%%%%%%%%%%%

%%% Use this command to include your bibliography file.
\newpage
\bibliography{m1732}

%%%%%%%%%%%%%%%%%%%%%%%%%%%%%%%%%%%%%%%%%%%%%%%%%%%%%%%%%%%%%%%%%%%%%%

\appendix
\clearpage
\section{Additional Datasets and Their Results} \label{additional_datasets}
\subsection{Additional Datasets and Experimental Setup}
\begin{table}[ht]
        \begin{center}
        \caption{Statistics of the two addtional datasets.}
        \vspace{0.2cm}
	%\small
        \resizebox{1.0\columnwidth}{!}{
	\begin{tabular}{lcccccc}
	\hline
	  & {Train} & {Dev} & {Test} & {\#C} & {|Q|} & {|C|}\\
	\hline
	PIQA & 14500 & 1613 & 1838 & 2 & 7.07 & 19.40\\
	SocialIQA & 31500 & 1910 & 1954 & 3 & 20.18 & 3.61\\
	\hline
	\end{tabular}}
 \label{tab:benchmarks_addtional_datasets}
         \end{center}
\end{table}

\begin{table}[ht]
\caption{The selection of the optimal learning rate and batch size.}
	\begin{center}
	%\small
	\begin{tabular}{lccc}
	\hline
	  Models & Hyperparameters & {PIQA} & {SocialIQA} \\
	\hline
	\multirow{2}{*}{{T5}} & Learning Rate & 1e-4 & 5e-5\\
         & Batch Size & 8 & 16\\
         \hline
	\multirow{2}{*}{{Unified-T5}} & Learning Rate & 1e-4 & 5e-5\\
         & Batch Size & 16 & 16\\
	\hline
	\end{tabular}
 \label{tab:parameters_addtional_datasets}
 \end{center}
\end{table}
In addition to the five benchmarks discussed in Section \ref{Datasets}, we also conducted experiments on two additional MCQA datasets. The Physical Interaction Question Answering (PIQA) dataset \citep{bisk2020piqa} is designed for commonsense reasoning and was created to evaluate the physical knowledge of existing NLP models. The Social Interaction QA (SocialIQA) dataset \citep{sap2019social} is a question-answering benchmark that tests social commonsense intelligence, assessing models' abilities to reason about the social implications of everyday events and situations. The statistics for these two additional benchmarks are provided in Table \ref{tab:benchmarks_addtional_datasets}.

Following the procedure outlined in Section \ref{Experimental Setup}, we conducted a search over the ranges \{1e-4, 5e-5, 1e-5, 5e-6\} and \{16, 8, 4\} to determine the optimal learning rate and batch size, respectively. The final selections are presented in Table \ref{tab:parameters_addtional_datasets}. Then, for the two datasets, we performed multiple experiments with different random seeds \{1,10,20\} and reported the mean and standard deviation.

\subsection{Additional Results and Analysis}
\begin{table}[ht]
    \begin{center}
    \caption{Comparison on the two additional MCQA benchmarks (PIQA, SocialIQA). All results are from our evaluations.}
    \vspace{0.2cm}
	%\small
    \resizebox{1.0\columnwidth}{!}{
	\begin{tabular}{lcccc}
	\hline
	\multirow{2}{*}{{Model}} & \multicolumn{2}{c}{{PIQA}}  & \multicolumn{2}{c}{{SocialIQA}}\\
	& {Dev} & {Test} & {Dev} & {Test}\\
	\hline
	T5-Base\\
        \quad T5-vanilla & 66.79($\pm$0.18) & 64.96($\pm$0.61) & 75.73($\pm$0.29) & 63.27($\pm$0.44)\\
        \quad GenMC\textsubscript{T5} & 71.75($\pm$0.35) & 71.39($\pm$0.25) & \textbf{79.13($\pm$0.20)} & \textbf{66.91($\pm$0.31)}\\
	\quad Our Model & \textbf{72.22($\pm$0.58)} & \textbf{72.42($\pm$0.62)} & 78.97($\pm$0.20) & 66.73($\pm$0.55)\\
        \hline
        U-T5-Base\\
	\quad UnifiedQA\textsubscript{T5} & 26.66($\pm$0.00) & 25.73($\pm$0.00) & 55.29($\pm$0.00) & 45.80($\pm$0.00)\\
	\quad UnifiedQA\textsubscript{T5-FT} & 70.78($\pm$0.16) & 70.22($\pm$0.56) & 78.53($\pm$0.38) & 66.46($\pm$0.59)\\
        \quad Our Model & \textbf{72.70($\pm$0.18)} & \textbf{72.27($\pm$0.79)} & \textbf{79.08($\pm$0.37)} & \textbf{66.70($\pm$0.21)}\\
	\hline
	T5-Large\\
        \quad T5-vanilla & 69.85($\pm$1.06) & 69.03($\pm$1.60) & 79.35($\pm$0.11) & 69.77($\pm$0.32)\\
        \quad GenMC\textsubscript{T5} & \textbf{79.36($\pm$0.51)} & \textbf{78.84($\pm$0.36)} & 82.02($\pm$0.16) & 73.01($\pm$0.52)\\
	\quad Our Model & 78.47($\pm$0.13) & 78.25($\pm$0.98) & \textbf{82.15($\pm$0.30)} & \textbf{74.07($\pm$0.23)}\\
        \hline
	U-T5-Large\\
        \quad UnifiedQA\textsubscript{T5} & 31.12($\pm$0.00) & 31.01($\pm$0.00) & 60.37($\pm$0.00) & 51.69($\pm$0.00)\\
	\quad UnifiedQA\textsubscript{T5-FT} & \textbf{79.60($\pm$0.23)} & \textbf{78.53($\pm$0.41)} & 81.52($\pm$0.09) & 73.39($\pm$0.29)\\
	\quad Our Model & 79.09($\pm$0.36) & 77.98($\pm$0.33) & \textbf{81.96($\pm$0.17)} & \textbf{74.31($\pm$0.64)}\\
	\hline
	\end{tabular}
    }
    \label{tab:main_result_addtional_datasets}
    \end{center}
\end{table}

Table \ref{tab:main_result_addtional_datasets} presents the comparative results on the two additional datasets. The experimental findings indicate that our model outperforms most baseline models on these datasets, further validating the effectiveness of our reasoning approach. Notably, for the PIQA dataset, where the questions are short and the choices are long, capturing the tokens of the question attended to by all the choices (commonalities) from those by individual choices (nuances) is inherently more challenging. Nevertheless, our model performs comparably to the baselines and even surpasses them under the T5-Base and Unified-T5-Base setup.

%%%%%%%%%%%%%%%%%%%%%%%%%%%%%%%%%%%%%%%%%%%%%%%%%%%%%%%%%%%%%%%%%%%%%%

\end{document}